\documentclass[letterpaper]{article} 
\usepackage[submission]{aaai24}  
\usepackage{times}  
\usepackage{helvet}  
\usepackage{courier}  
\usepackage[hyphens]{url}  
\usepackage{graphicx} 
\usepackage{natbib}  
\usepackage{caption} 
\usepackage[algo2e, ruled, vlined]{algorithm2e}
\usepackage{algorithm}

\usepackage{newfloat}
\usepackage{listings}
\DeclareCaptionStyle{ruled}{labelfont=normalfont,labelsep=colon,strut=off} 
\floatstyle{ruled}
\newfloat{listing}{tb}{lst}{}
\floatname{listing}{Listing}
\usepackage{booktabs}
\usepackage{amsmath}
\usepackage{amssymb}
\usepackage{mathtools}
\usepackage{amsthm}
\usepackage{bm}
\usepackage{multirow}
\usepackage{bibentry}

\usepackage[capitalise]{cleveref}
\usepackage{graphicx}

\DeclareMathOperator*{\argmin}{arg\,min}
\newcommand*{\argmina}{\argmin\limits}

\title{Semi-Supervised Learning of Dynamical Systems with Neural Ordinary Differential Equations: A Teacher-Student Model Approach}
\author{
    Yu Wang\textsuperscript{\rm 1},
    Yuxuan Yin\textsuperscript{\rm 1},
    Karthik Somayaji Nanjangud Suryanarayana\textsuperscript{\rm 1},
    J\'an Drgo\v na\textsuperscript{\rm 2},
    Malachi Schram\textsuperscript{\rm 4},
    Mahantesh Halappanavar\textsuperscript{\rm 2},
    Frank Liu\textsuperscript{\rm 5},
    Peng Li\textsuperscript{\rm 1}
}
\affiliations{
    \textsuperscript{\rm 1} University of California, Santa Barbara\\
    \textsuperscript{\rm 2} Pacific Northwest National Laboratory\\
    \textsuperscript{\rm 4} Thomas Jefferson National Accelerator Facility\\
    \textsuperscript{\rm 5} Oak Ridge National Laboratory\\

    \{yu95, karthi, y\_yin, lip\}@ucsb.edu, jan.drgona@pnnl.gov, schram@jlab.org, liufy@ornl.gov
}

\begin{document}

\maketitle

\begin{abstract}
Modeling dynamical systems is crucial for a wide range of tasks, but it remains challenging due to complex nonlinear dynamics, limited observations, or lack of prior knowledge. Recently, data-driven approaches such as Neural Ordinary Differential Equations (NODE) have shown promising results by leveraging the expressive power of neural networks to model unknown dynamics. However, these approaches often suffer from limited labeled training data, leading to poor generalization and suboptimal predictions. On the other hand, semi-supervised algorithms can utilize abundant unlabeled data and have demonstrated  good performance in classification and regression tasks.
We propose TS-NODE, the first semi-supervised approach to modeling dynamical systems with NODE. TS-NODE explores cheaply generated synthetic pseudo rollouts to broaden exploration in the state space and to tackle the challenges brought by lack of ground-truth system data under a teacher-student model.  TS-NODE employs an unified optimization framework that corrects the teacher model based on the student's feedback while mitigating the potential false system dynamics present in pseudo rollouts.
TS-NODE demonstrates significant performance improvements over a baseline Neural ODE model on multiple dynamical system modeling tasks.


\end{abstract}

\begin{figure*}[htb!]
    \centering
    \includegraphics[width=1.0\textwidth]{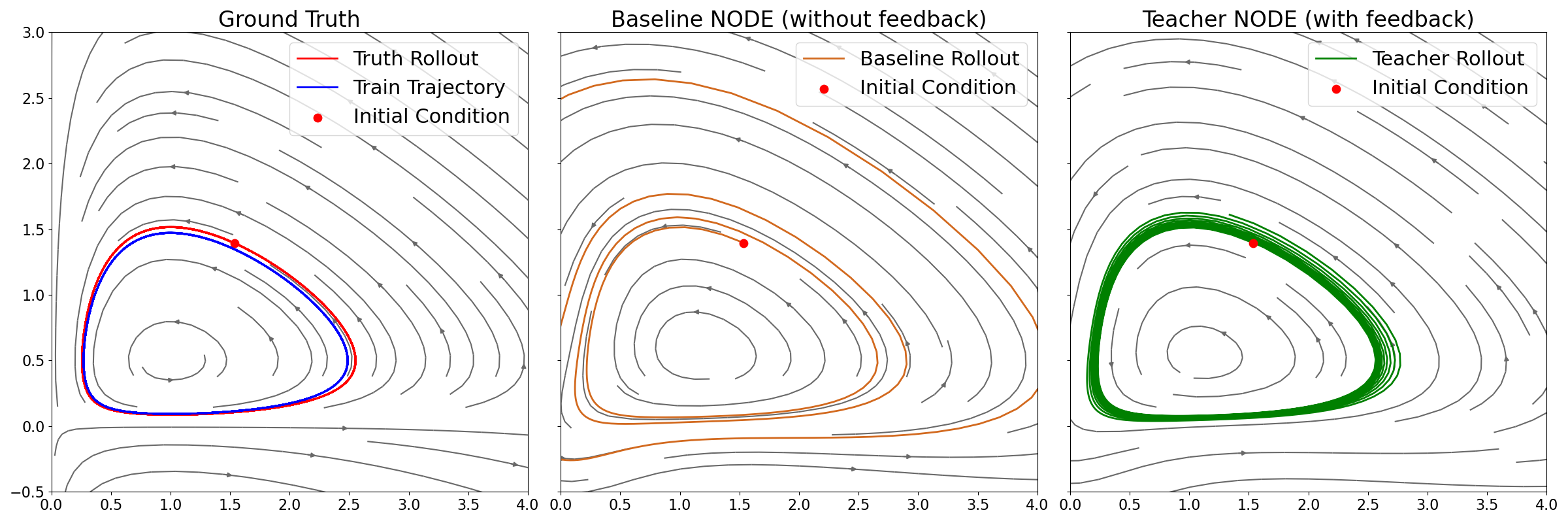}
    \caption{Phase portraits and rollouts on a test initial condition of different learned models in the phase space of a Lotka-Volterra system\cite{lk}. Left: The ground truth. Middle: the baseline neural ODE. Right: the proposed teacher model with student feedback. All models are trained with a single train trajectory (blue line in the left). }
    \label{fig:demonstration_lk}
\end{figure*}

\section{Introduction}
Precise modeling of system dynamics plays a pivotal role across multiple domains, including system identification, model predictive control, and reinforcement learning. Nonetheless, real-world systems frequently exhibit a high degree of non-linearity within a high-dimensional state space, posing considerable challenges in modeling dynamics without adequate prior knowledge.

Recent advancements in machine learning enable the use of data-driven methods to directly model system dynamics from observations \cite{hnn, gp_2015, koopman_2018, node}, effectively circumventing the challenges associated with making appropriate prior model assumptions. For instance, \cite{gp_2015} employs a Gaussian Process to model dynamical systems, providing a data-efficient and probabilistic method for learning in the presence of uncertainty and noise. Other studies, such as  \cite{koopman_2018, koopman_2020}, utilize deep neural networks to learn the Koopman operator, which represents a linear embedding of non-linear dynamical systems. Furthermore, Neural Ordinary Differential Equations (NODE)\cite{node} leverages a black-box neural network to model dynamics as an ordinary differential equation (ODE), demonstrating significant potential in modeling unknown dynamical systems.

However, one significant obstacle for data-driven methods stems from the practical difficulty in acquiring sufficient data for modeling dynamical systems. The data collection process can be expensive, time-consuming, and in some cases, even infeasible due to physical constraints or safety considerations. The shortage of data results in limited coverage of the state space, leading to gaps in the model's understanding and poor prediction ability.

Semi-supervised learning methods have emerged as a promising solution to the problem of limited labeled data in classification and regression tasks. These approaches effectively leverage  a large amount of unlabeled data along with some labeled data. Unlabeled data can  be either obtained directly from observations with unknown or uncollected labels or generated through various data augmentation techniques \cite{simclr, Cubuk_2019_CVPR, NEURIPS2020_d85b63ef}. Semi-supervised methods typically assume that unlabeled data share the same structural information as labeled data and often utilize them in different ways, such as self-training \cite{meta_pseudo_label}, multi-view training \cite{simclr, wang2020hypersphere}, or based on graph methods \cite{semi_graph_2017, semi_graph_2019}. These approaches promote the model to produce similar outputs for similar instances or preserve the structure of the data in the learned representation.

Thus, it is logical to consider the integration of semi-supervised learning into the modeling of dynamical systems with neural ODEs, a concept yet to be explored by the research community. However, the adaptation of semi-supervised learning strategies to the problem is not straightforward and presents several challenges: the \emph{definition, generation} and \emph{utilization} of unlabeled data in dynamical system modeling significantly differ from their counterparts in other semi-supervised learning contexts such as classification and regression. For instance, generating additional dynamical training data by directly applying popular  data augmentation and multi-view learning techniques may be detrimental, as alternating truthfully observed system data can inject false system dynamics.

To address these challenges, we introduce TS-NODE, a semi-supervised learning approach designed specifically for modeling dynamical systems with neural ODEs. Our contributions are:
\begin{itemize}
    \item We present the first teacher-student model based approach to enabling semi-supervised modeling of dynamical systems; 
    
    \item We explore synthetic \emph{pseudo rollouts}  generated by the teacher model to broaden exploration in the state space and to tackle the challenges brought by lack of ground-truth system data; 

    \item We employ an unified teacher-student optimization framework that corrects the teacher model based on the student's feedback while mitigating the potential false system dynamics present in pseudo trajectories;
    
    \item We evaluate TS-NODE  on multiple dynamical system modeling tasks, demonstrating significant performance improvements over a baseline Neural ODE model.
\end{itemize}

\section{Preliminaries}
\subsection{Neural ODE for Modeling Dynamical Systems}
A neural ODE parameterized by $\bm{\theta}$ models the dynamics of the evolution of the state  ${\bm{y}}(t)$ of an unknown system with a neural network\cite{node}:
\begin{equation}
    \dot{\bm{y}}(t) = \text{NN}(\bm{y}(t), t;\ \bm{\theta}).
\end{equation}

Let $\bm{y}=\left\{\bm{y}(t_{0}),\bm{y}(t_{1})\cdots \bm{y}(t_{n})\right\}$ be an observed system trajectory in the state space over a period of time and denote its initial condition\footnote{In later sections, $\bm{y}$ refers to a time series and $\bm{y_{0}}$ represents its value at the first time point unless stated otherwise.} by $\bm{y_{0}}$. A neural ODE is trained to learn the system dynamics by matching its predicted trajectory (rollout) with the true trajectory. 
The neural ODE's predicted trajectory is obtained by solving an initial value problem (IVP) for each $t_{i},  i=[0, 1\cdots n]$:
\begin{equation}
    \hat{\bm{y}}(t_{i}) = \bm{y}(t_{0}) + \int_{t_{0}}^{t_{i}} \text{NN}(\hat{\bm{y}}(\tau), \tau;\ \bm{\theta}) d\tau
\end{equation}

The  parameters  $\bm{\theta}$ are optimized by minimizing the mean squared error (MSE) between $\hat{\bm{y}}$ and $\bm{y}$:
\begin{equation}
    \bm{\theta^{*}}= \argmina_{\bm{\theta}}\mathcal{L}(\bm{y};\bm{\theta}) = \argmina_{\bm{\theta}} \text{MSE}(\bm{y}, \hat{\bm{y}}(\bm{\theta})). 
\end{equation}

\subsection{Challenges in Semi-supervised Modeling of Dynamical Systems with Neural ODEs}\label{sec:challenges}
Training neural ODEs 
requires a substantial amount of data for optimal generalization. On the other hand, data for certain systems like a complex control system can be costly to collect and the observed trajectories often have limited coverage of the state space. Scarcity of data results in poor generalization and suboptimal long-horizon rollouts of the trained neural ODE models.  
The ability of making use of large amounts of cheap unlabeled data makes semi-supervised learning particularly appealing for dynamical system modeling. 
Nevertheless, semi-supervised learning has not been attempted for neural ODE based dynamical system modeling before. It is not immediately evident how to pursue semi-supervised learning under this new context, raising several crucial questions we outline below:  

\textbf{Q1: What does  \emph{unlabeled data} mean for dynamical system modeling?}
The concept of unlabeled data in classical semi-supervised learning\cite{semi_review1, semi_review2}, e.g. for classification and regression tasks is well defined:  unlabeled data are input examples for which there exist no ground-truth output label. However, when it comes to modeling dynamics with Neural ODEs, this definition becomes less clear as the system model is trained on observed system trajectories rather than a set of input/output pairs with or without the output label.

\textbf{Q2: How to generate \emph{useful} unlabeled data for dynamical system modeling?}
Data augmentation has been popular for introducing new training experiences without added data labeling effort in many learning tasks.  In most of cases, it works on the premise that subtle input perturbations do not change or substantially alter the output labels. A brute-force application of existing data augmentation techniques such as crop-and-resize\cite{simclr} or random erasing\cite{zhong2017random} to dynamic system modeling may not be completely appropriate: these augmentation transformations either cannot directly operate on data that are dynamical in nature, or fail to introduce meaningful new experiences that are consistent with the underlying dynamics to be modeled, leading to degraded learning performance.  


\textbf{Q3: How to mitigate \emph{misinformation} in the unlabeled data?}
Any generated unlabeled data are inevitable to exhibit dynamical behavior that deviates from the system under modeling. It is imperative to prevent the false dynamical information from such data to leak into the system model.

\section{TS-NODE Framework Overview} 

We argue that the proposed teacher-student model based TS-NODE approach is well positioned to address the issues raised in aforementioned the three questions.

First, under our context data that are useful must exhibit dynamical behavior relevant to the task of dynamical system learning. As such, meaningful ``unlabeled data'' can be considered to be state trajectories that do not necessarily fully reflect but bear resemblance  to the underlying true system dynamics. We call such trajectories \emph{pseudo rollouts}.

Second, to best aid dynamical system modeling, the employed pseudo rollouts shall comply with the typical physical constraints of the family of  systems to which the targeted system belongs. In other words, augmenting known system trajectories via unconstrained non-physical transformations provides little value if it does not hamper system modeling. To this end, it is most straightforward to generate pseudo rollouts, a.k.a. unlabeled data, by a dynamical system model that is similar to the system under modeling.

\begin{figure}[ht!]
    \centering
    \includegraphics[width=0.45\textwidth]{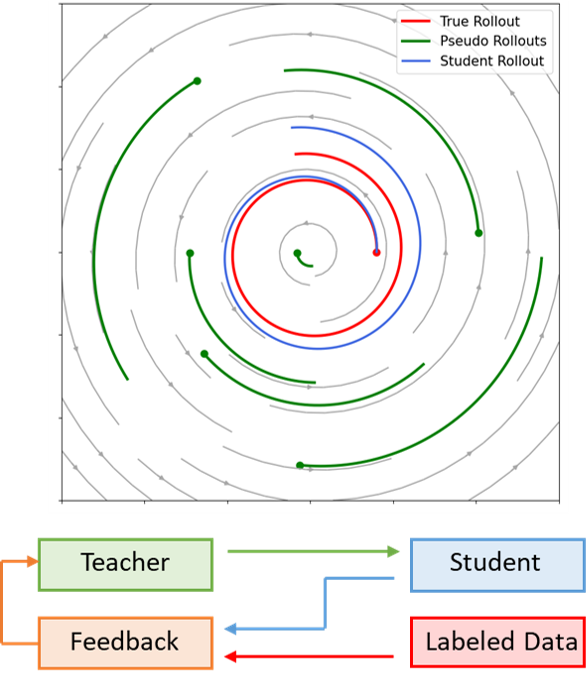}
    \caption{An illustrated overview of the learning mechanism of TS-NODE. The teacher model generates pseudo rollouts (green lines) with more coverage of the state space. The student model learns from those trajectories and then predict rollout (blue line) with the initial condition of the true trajectory (red line). The MSE between student's prediction and the ground truth serves as the feedback signal to refine the teacher's prediction.}
    \label{fig:overview}
\end{figure}

TS-NODE achieves this by making use of a teacher neural ODE model, which is trained with (limited) true system trajectories, to generate  pseudo rollouts, potentially covering part of the state space with little or no true system dynamical data. The generated trajectories, or pseudo rollouts, encapsulate the teacher's present belief about the underlying dynamics and provide contextual awareness of the target problem. Once generated, they can be readily used to train the student, another neural ODE.

Lastly, directly utilizing the pseudo rollouts in the final system model building  can lead to undesired amplification of inaccurate knowledge. Instead, TS-NODE utilizes these trajectories to train the student model whose performance on a validation dataset acts as a feedback to the teacher to correct the teacher's belief.   


\subsubsection{Basic Design of TS-NODE Model}
With the same high-level spirit of the semi-supervised learning technique for image classification \cite{meta_pseudo_label}, TS-NODE jointly optimizes the teacher and student model at each training step. First, pseudo rollouts starting from a set of initial conditions in the state space are generated by the teacher, which are used to train the student neural ODE whose performance is evaluated on the true system trajectories. The evaluation loss of the student, called the student feedback loss, is considered as a feedback to the teacher as it depends on the teacher's pseudo rollouts. The teacher is then updated by minimizing the student feedback loss and its direct fitting loss over the limited true system trajectories, a.k.a. the labeled dataset.


\subsection{Basic Learning Flow of TS-NODE}
Following the standard definitions in semi-supervised learning, we denote the labeled data, i.e. true system dynamical trajectories by: $\mathcal{D}^{l}:=\left\{\bm{y^{l}}\right\}$. We denote the unlabeled data by the pseudo rollouts predicted from the teacher: $\mathcal{D}^{u}:=\left\{\bm{y^{u}}\right\}$, with their initial conditions denoted by $\bm{y_{0}^{u}}$.
Two NODEs are introduced in TS-NODE, namely the teacher: $T(\cdot;\bm{\theta_{T}})$ and the student: $S(\cdot;\bm{\theta_{S}})$. 

The training of TS-NODE involves four major steps:
\begin{itemize}
    \item [1.] Generate pseudo rollouts from the teacher: \\
    \begin{equation}
        \bm{y_{T}^{u}}(\bm{\theta_{T}})=T(\bm{y_{0}^{u}};\bm{\theta_{T}})
        \label{eq:4}
    \end{equation}
    \item [2.] Train the student\footnote{We follow the suggestion in \cite{meta_pseudo_label} to update the student with one-step gradient descent.} with an unlabeled loss ($\mathcal{L}_{S}^{u}$) defined on pseudo rollouts: \\
    \begin{equation}
        \mathcal{L}_{S}^{u} = \text{MSE}\left(\bm{y_{T}^{u}}(\bm{\theta_{T}}),\ S(\bm{y_{0}^{u}},\bm{\theta_{S}})\right)
    \end{equation}
    \begin{equation}
        \label{eq:s_update}
        \bm{\theta_{S}^{'}} = \bm{\theta_{S}} - \eta_{S}\cdot \nabla_{\bm{\theta_{S}}}\cdot \mathcal{L}_{S}^{u}
    \end{equation}
    \item [3.] Calculate the error of the updated student model on $\mathcal{D}^{l}$ as the feedback loss $\mathcal{L}_{F}^{l}$: 
    \begin{equation}
        \mathcal{L}_{F}^{l} := \mathcal{L}_{S^{'}}^{l}= \text{MSE}(\mathcal{D}^{l}, S^{'}(\bm{y_{0}^{l}};\bm{\theta_{S}^{'}}(\bm{\theta_{T}}))))
        \label{eq:7}
    \end{equation}
    \item [4.] Train the teacher with both the labeled loss $\mathcal{L}_{T}^{l}$ and the feedback loss $\mathcal{L}_{F}^{l}$:
    \begin{equation}
        \mathcal{L}_{T}^{l}=\text{MSE}(\mathcal{D}^{l}, T(\bm{y_{0}^{l}};\bm{\theta_{T}}))
    \end{equation}
    \begin{equation}
        \bm{\theta_{T}^{'}} = \bm{\theta_{T}} - \eta_{T}\cdot \nabla_{\bm{\theta_{T}}}\cdot (\mathcal{L}_{T}^{l} + \mathcal{L}_{F}^{l})
        \label{eq:9}
    \end{equation}
\end{itemize}

\subsection{Gradient for Updating the Teacher and Its Scalable Computation}\label{sec:gradient}
To update the teacher with the feedback loss involves computing a Jacobian matrix or calculating per-sample gradients, which can result in a large computational graph and computational inefficiency (More details in Appendix A). Instead, we make a modification to the framework by employing a noisy teacher to generate the pseudo rollouts. Utilizing a noisy teacher enables the derivation of scalable gradients using the REINFORCE rule\cite{reinforce} and more interpretable, alternative feedback loss.
\paragraph{Noisy Teacher:}
We design a simple noisy teacher $T_{\epsilon}$ by adding independent constant noise to each step of the rollout of the standard teacher. The rollouts of $T_{\epsilon}$ are thus samples from a Gaussian distribution:
\begin{equation}
    \bm{y_{T_{\epsilon}}}^{n\times d} \sim \mathcal{N}(\bm{y}_{T}^{n\times d};\sigma\cdot \bm{I}^{(n\times d)\times (n\times d)};\bm{\theta_{T}})
\end{equation}
where $n$ is the number of observed time steps, $d$ is the dimension of the state space.
\paragraph{Derivation of the Feedback Gradient:}
For a noisy teacher, since the pseudo rollouts are samples from a distribution, \cref{eq:s_update} needs to be rewritten as:
\begin{equation}
    \label{eq:s_udpate_exp}
    \bm{\theta_{S}^{'}} = \bm{\theta_{S}} - \eta_{S}\cdot \mathbb{E}_{\bm{y}_{T_{\epsilon}}^{u}\sim p(\bm{y}_{T_{\epsilon}}^{u})}\left[\nabla_{\bm{\theta_{S}}}\cdot \mathcal{L}_{S}^{u}\right]
\end{equation}

For convenience, we write the gradient to update the student as\ \   $\bm{g_{S}}:=\nabla_{\bm{\theta_{S}}}\cdot \mathcal{L}_{S}^{u}$, the gradient of $\mathcal{L}_{F}^{l}$ with respect to $\bm{\theta_{T}}$ as\ \ $\bm{g_{F}}:=\nabla_{\bm{\theta_{T}}}\cdot \mathcal{L}_{F}^{l}$. By the chain rule
, we have:
\begin{equation}
\begin{aligned}
    \left[\bm{g_{F}}\right]^{T}=\frac{\partial\mathcal{L}_{F}^{l}}{\partial\bm{\theta_{T}}}=\frac{\partial\mathcal{L}_{F}^{l}}{\partial\bm{\theta_{S}^{'}}}\cdot \frac{\partial\bm{\theta_{S}^{'}}}{\partial\bm{\theta_{T}}}
\end{aligned}
\end{equation}

By substituting the $\bm{\theta_{S}^{'}}$ in \cref{eq:s_udpate_exp}, we have:
\begin{equation}
    \left[\bm{g_{F}}\right]^{T}=-\eta_{S}\cdot \frac{\partial\mathcal{L}_{F}^{l}}{\partial\bm{\theta_{S}^{'}}}\cdot \underbrace{{\frac{\partial}{\partial\bm{\theta_{T}}}\left(\mathbb{E}_{\bm{y_{T_{\epsilon}}^{u}}\sim p(\bm{y_{T_{\epsilon}}^{u}})}\left[\bm{g_{S}}\right]\right)}}_{\text{jacobian}}
\end{equation}

By REINFORCE rule\cite{reinforce}, we rewrite the Jacobian as :
\begin{equation}
    \begin{aligned}
        \frac{\partial}{\partial\bm{\theta_{T}}}\left(\mathbb{E}_{\bm{y_{T_{\epsilon}}^{u}}\sim p(\bm{y_{T_{\epsilon}}^{u}})}\left[\bm{g_{S}}\right]\right)=\bm{g}_{S}\cdot \frac{\partial\log p(\bm{y_{T_{\epsilon}}^{u}})}{\partial\bm{\theta_{T}}}
    \end{aligned}
\end{equation}

Finally, the feedback gradient is derived as:
\begin{equation}
    \label{eq:g_f}
    \left[\bm{g_{F}}\right]^{T}=-\eta_{S}\cdot \underbrace{\vphantom{\frac{\partial\mathcal{L}_{F}^{l}}{\partial\bm{\theta_{S}^{'}}}} \frac{\partial\mathcal{L}_{F}^{l}}{\partial\bm{\theta_{S}^{'}}}}_{a}\cdot \underbrace{\vphantom{\frac{\partial\mathcal{L}_{F}^{l}}{\partial\bm{\theta_{S}^{'}}}} \bm{g_{S}}}_{b}\cdot \underbrace{\vphantom{\frac{\partial\mathcal{L}_{F}^{l}}{\partial\bm{\theta_{S}^{'}}}} \frac{\partial\log p(\bm{y_{T_{\epsilon}}^{u}})}{\partial\bm{\theta_{T}}}}_{c}
\end{equation}

In \cref{eq:g_f}, all three parts can be calculated efficiently with one step of automatic differentiation.

\paragraph{Approximated Alternative Feedback Loss:}
We show that \cref{eq:g_f} can be further simplified by utilizing Taylor expansion. An approximated alternate feedback loss can be derived which can give us a direct interpretation of how the feedback loss helps to refine teacher's pseudo rollouts.

First, we rewrite $\bm{g_{F}}$ by applying Taylor expansion to $\mathcal{L}_{F}^{l}(\bm{\theta_{S}^{'}})$ and ignoring higher order terms:
\begin{equation}
    \label{eq:taylor_exp}
    \mathcal{L}_{F}^{l}(\bm{\theta_{S}^{'}})\approx \mathcal{L}_{F}^{l}(\bm{\theta_{S}}) + \frac{\partial\mathcal{L}_{F}^{l}(\bm{\theta_{S}^{'}})}{\partial\bm{\theta_{S}^{'}}}\cdot (\bm{\theta_{S}^{'}}-\bm{\theta_{S}})
\end{equation}

Combining \cref{eq:taylor_exp} and \cref{eq:s_update} we have:
\begin{equation}
    \begin{aligned}
        \mathcal{L}_{F}^{l}(\bm{\theta_{S}^{'}})-\mathcal{L}_{F}^{l}(\bm{\theta_{S}})&\approx \frac{\partial\mathcal{L}_{F}^{l}(\bm{\theta_{S}^{'}})}{\partial\bm{\theta_{S}^{'}}}\cdot (\bm{\theta_{S}^{'}} - \bm{\theta_{S}})\\
        &= \frac{\partial\mathcal{L}_{F}^{l}(\bm{\theta_{S}^{'}})}{\partial\bm{\theta_{S}^{'}}}\cdot (-\eta_{S}\cdot \bm{g_{S}})
    \end{aligned}
\end{equation}

Thus, the feedback gradient can be approximated as: 
\begin{equation}
    \label{eq:g_feedback}
    \begin{aligned}
         \left[\bm{g_{F}}\right]^{T}&\approx \left(\mathcal{L}_{F}^{l}(\bm{\theta_{S}^{'}})-\mathcal{L}_{F}^{l}(\bm{\theta_{S}})\right)\cdot \frac{\partial\log p(\bm{y}_{T_{\epsilon}}^{u})}{\partial\bm{\theta_{T}}}\\
         &=\left(\mathcal{L}_{S^{'}}^{l}-\mathcal{L}_{S}^{l}\right)\cdot \frac{\partial\log p(\bm{y}_{T_{\epsilon}}^{u})}{\partial\bm{\theta_{T}}}
    \end{aligned}
\end{equation}

Note that the feedback gradient in \cref{eq:g_feedback} corresponds to the gradient of a scaled negative log-likelihood (NLL) loss and we can rewrite the original feedback loss as:
\begin{equation}
    \mathcal{L}_{F}^{l}\approx \underbrace{\left(\mathcal{L}_{S}^{l}-\mathcal{L}_{S^{'}}^{l}\right)}_{h:\text{student's improvement}}\cdot\quad  \text{NLL}(\bm{y}_{T_{\epsilon}}^{u};\bm{\theta_{T}})
    \label{eq:19}
\end{equation}

The alternate feedback loss refines the noisy teacher by modulating the likelihood of generating particular pseudo rollouts based on the student's improvement. After trained with given a rollout, if the student's performance sees an enhancement (i.e., $h \ge 0$), minimizing $\mathcal{L}_{F}^{l}$ effectively maximizes the log-likelihood of that rollout, thereby augmenting the probability that the teacher will generate similar rollouts in future iterations. Conversely, if the student's performance deteriorates, the feedback mechanism will work to diminish the likelihood of generating such rollouts.

In our experiments, optimization of TS-NODE follows the procedure from \cref{eq:4} to \cref{eq:9}. The original feedback loss in \cref{eq:7} is replaced with the one defined in \cref{eq:19} for efficient computation. We summarize the pseudo code for TS-NODE in Appendix B for further reference.

\section{Experiments}
We evaluate TS-NODE on several aspects: 1) We compare TS-NODE with the baseline NODE to validate its overall efficacy; 2) We show TS-NODE provides a better way of generating and utilizing unlabeled data by compared with NODE trained with basic data augmentations. 3) We compare TS-NODE with a variant that disables the student's feedback to underscore the necessity of the feedback loop.

\subsection{Datasets and Model Configuration} 
We adopt three dynamical systems, the converging cubic system \cite{node}, Lotka-volterra system \cite{lk} and a simple pendulum, which are commonly in the literature, as the test cases. We simulate each ground-truth system to generate a single state trajectory from a predetermined initial condition as the only labeled training data, to evaluate the baseline \cite{node} and proposed TS-NODE under limited observation.
To evaluate the generalization of each learned model, we randomly select a set of 20 initial conditions in the state space, and simulate the ground-truth model to obtain 20 state trajectories as the testing data, as detailed next.
\paragraph{Converging Cubic System:} The converging cubic system we employed is a 2-dimensional system, adapted from \cite{node}. The system dynamics is described as below:
\begin{equation}
\begin{aligned}
    \dot{x} = a\cdot x^{3} + b\cdot y^{3}\\
    \dot{y} = c\cdot x^{3} + d\cdot y^{3}
\end{aligned}
\end{equation}
We set $[a, b, c, d] = [0.1, 2, -2, -0.1]$. The training trajectory is a rollout from $[x_{0}, y_{0}]=[3, -1]$. The testing initial conditions are sampled from $\mathcal{N}([x_{0}, y_{0}];0.3\cdot\bm{I}^{2\times2})$. The simulation time is 10s.

\paragraph{Lotka-Volterra System :} The Lotka-Volterra system\cite{lk}, or the predator-prey system, is dictated by the dynamics as:
\begin{equation}
\begin{aligned}
    &\dot{x} = \alpha\cdot x - \beta\cdot xy\\
    &\dot{y} = \delta\cdot xy -\gamma\cdot y
\end{aligned}
\end{equation}
We set $[\alpha, \beta, \delta, \gamma]=[\frac{2}{3}, \frac{4}{3}, 1, 1]$ . The training trajectory is a rollout from $[x_{0}, y_{0}]=[1.4, 1.4]$ and the testing initial conditions are sampled from $\mathcal{N}([x_{0}, y_{0}];0.3\cdot\bm{I}^{2\times2})$. The simulation time is 10s.

\paragraph{Simple Pendulum:} A simple Pendulum is modeled as: 
\begin{equation}
    \ddot{\theta} = -a\cdot \dot{\theta} - b\cdot \sin{\theta}
\end{equation}
Simple pendulumn can be fully described by the angular position $\theta$ and the angular speed $\dot{\theta}:=\omega$.
In our experiments, we set $[a, b]=[0, 1]$, (i.e., simulating an ideal pendulum). The training trajectory is a rollout from$[\theta_{0}, \omega_{0}]=[2, 0]$. The initial conditions for generating the testing data are sampled from $\mathcal{N}([\theta_{0}, \omega_{0}];0.3\cdot\bm{I}^{2\times2})$. The simulation time is 10s.

The length of training or testing trajectories is 1,000 time steps.  We split the training trajectory as batches of short observations (10 steps) for efficiency as in \cite{node}. The initial conditions used by the teacher to generate pseudo rollouts are sampled around the training trajectory in each iteration.

\subsection{Baseline and Evaluation Metrics:} We compare TS-NODE with the standard neural ODE in \cite{node}, as the baseline. Neural networks in the baseline and TS-NODE share the identical architecture.
We randomly select 20 initial conditions and simulate each ground-truth system to generate 20 1000-step full test trajectories.   

We adopt two evaluation metrics for our study. The first, referred to as \emph{local error}, assesses the quality of local dynamic modeling around the testing regions. 
We randomly sample parts of the 20 full-length testing trajectories to create a set of 1,000 short testing rollouts of 10 steps. 
The local error is defined as the sum of the MSE between these testing rollouts and the corresponding rollouts from each learned model starting from the same initial condition.

The \emph{rollouts error} metric, measures the models' ability to forecast over different lengths of time horizon. We let learned models generate rollouts from the initial condition of each pre-generated full testing trajectory for   5\%, 10\%, 20\%, 50\%, and 100\%, respectively of the full length (1,000 steps). The rollouts error of a particular length is computed as the average MSE between the ground truth model and each learned model over the 20 rollouts of that length. 

\paragraph{1) Comparing TS-NODE with baseline NODE:}
We train TS-NODE and the baseline neural ODE with Adam optimizer for 10,000 iterations. We record the error metrics of the two models once every 100 iterations.  \cref{tab:table_comparison} reports the performances of two models based on the averages of the last five recorded errors. 

\begin{table*}[htbp!]
    \small
    \centering
    \setlength{\tabcolsep}{5.6mm}{
    \begin{tabular}{cccccccc}
         \toprule
          \multirow{2}{*}{Dataset} & \multirow{2}{*}{Models} & \multirow{2}{*}{$\text{Local Error}^\downarrow$} & \multicolumn{5}{c}{$\text{Rollouts Error}^\downarrow$}\\
          &   &  &5\% & 10\% & 20\% & 50\% & 100\% \\
         \midrule
          \multirow{2}{*}{Cubic}& Baseline & 39.84 & 0.95 & 1.35 & 1.21 & 1.42 & 1.08 \\
          & TS-NODE & \textbf{32.77} & \textbf{0.87} & \textbf{1.21} & \textbf{1.07} & \textbf{1.12} & \textbf{0.96} \\
          \midrule
          \multirow{2}{*}{LK}& Baseline & 20.93 & 0.10 & 2.54 & 64.79 & 1200.50 & 7346.09 \\
          & TS-NODE & \textbf{12.84} & \textbf{0.02} & \textbf{0.14} & \textbf{0.55} & \textbf{20.58} & \textbf{244.88} \\
          \midrule
          \multirow{2}{*}{Pendulum} & Baseline & 40.49 & 0.06 & 0.20 & 0.51 & 5.77 & 507.17  \\
          & TS-NODE & \textbf{5.65} & \textbf{0.01} & \textbf{0.05} & \textbf{0.25} & \textbf{0.95} & \textbf{1.89} \\
          \bottomrule
    \end{tabular}
    }
    \caption{Local error and rollouts error (both lower the better) of the baseline NODE and the TS-NODE on three systems.}
    \label{tab:table_comparison}
\end{table*}

\begin{figure*}[!ht]
    \centering
    \includegraphics[width=0.99\textwidth]{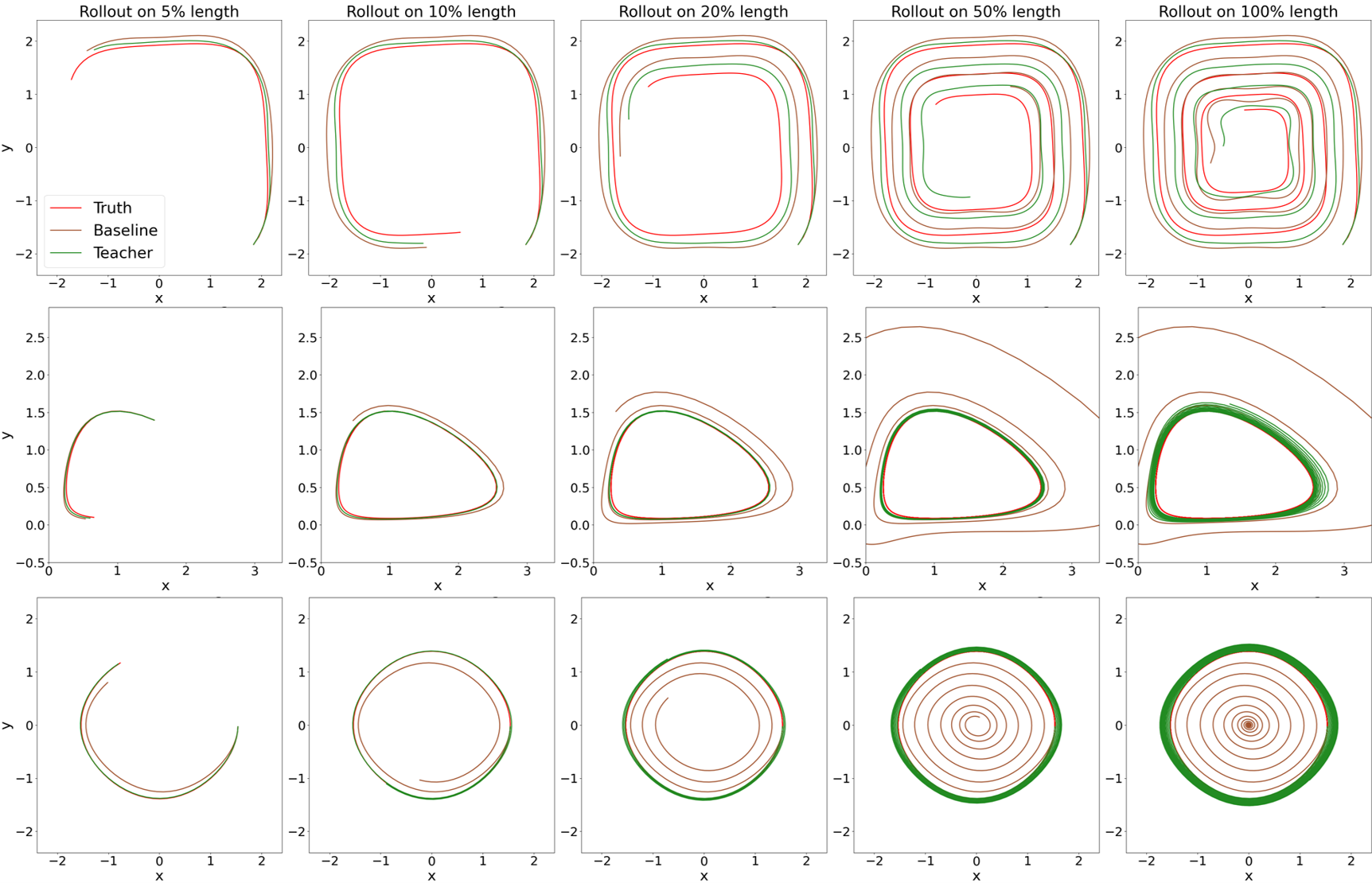}
    \caption{Rollouts from the trained baseline model and teacher model of TS-NODE in the phase space for three different systems. The rollouts are simulated from the initial condition of one pre-generated full testing trajectory for 5\%, 10\%, 20\%, 50\%, and 100\%, of the full length (1,000 steps). Top: converging cubic system. Middle: Lotka-Volterra system. Bottom: pendulum.}
    \label{fig:trace}
\end{figure*}

For all systems, TS-NODE consistently surpasses the baseline model across all metrics, particularly when making long-horizon predictions. The baseline model exhibits a significant error in generating extended rollouts for the LK system and the pendulum, suggesting its inability to accurately generalize the precise dynamics of these systems, leading to accumulation and subsequent explosion of error over time.

To further illustrate this point, we provide visual representations of the rollouts of varying lengths, as predicted by both the final trained baseline model and the teacher model of TS-NODE, originating from a single test initial condition (see \cref{fig:trace}). It is evident  that even in scenarios where the baseline model fails in producing accurate predictions, TS-NODE remains capable of accurately capturing and replicating the inherent dynamics over an extended duration.

\begin{figure*}[!ht]
    \centering
    \includegraphics[width=1.\textwidth]{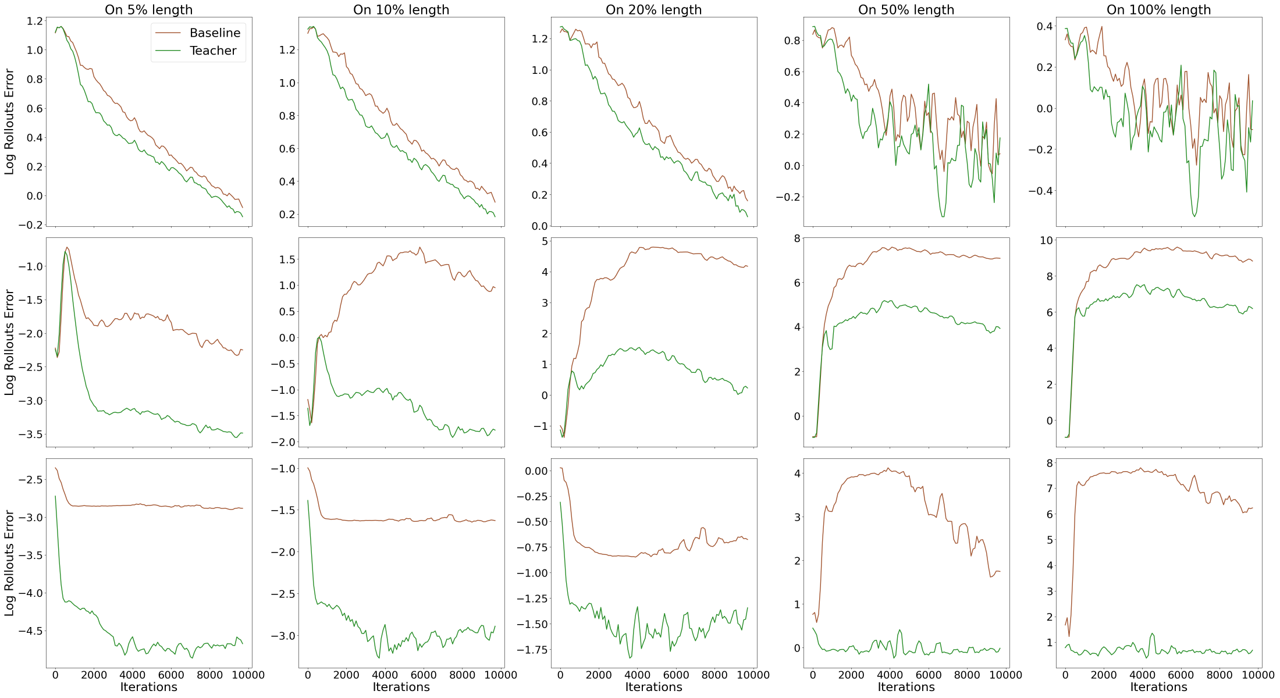}
    \caption{The rollouts errors of different length (5\%, 10\%, 20\%, 50\%, and 100\% of the full length (1000 steps)) of the baseline model and the teacher model. Top: converging cubic system. Middle: Lotka-Volterra system. Bottom: simple pendulum.}
    \label{fig:test_score}
\end{figure*}

We also plot the rollouts error during the training process for the baseline and the teacher in \cref{fig:test_score}. We take the logarithm of the rollouts error and apply a moving average to smooth the curve. The scarcity of the data results in a large rollout error of the baseline, which cannot be reduced or can even increase with extended training. In contrast, TS-NODE consistently shows better test performance across varying rollout lengths as the training proceeds, clearly demonstrating the overall effectiveness of TS-NODE framework. 

\paragraph{2) Comparing TS-NODE with NODEs trained by basic data augmentations:}
We investigated the effects of two common data augmentation techniques: the addition of white noise ($\sigma=0.01$) and rescaling by 0.95X. We apply them to the labeled dataset to generate an unlabeled dataset and combine both datasets to train the standard neural ODEs. We denote the models trained with different augmentations by "White Noise" and "Re-scale", respectively. Experiment results on the LK dataset are shown in \cref{tab:table_ablation}.

TS-NODE surpasses all models trained using basic data augmentations, which can potentially compromise the dataset's intrinsic physical properties. This performance underscores that TS-NODE offers a superior approach to generating and utilizing unlabeled data through its pseudo rollouts and the feedback loop.

\paragraph{3) The necessity of the feedback loop: } We disable the feedback loop by training the teacher only with labeled loss and maintaining the student's learning from teacher-generated pseudo rollouts. The performance of the final student is tabulated as "No Feedback" in \cref{tab:table_ablation}.

Without the feedback signal, the misinformation in the pseudo rollouts cannot get exposed via students evaluation, and the teacher cannot refine itself to generate better pseudo rollouts, resulting a bad-behaved final student. With the feedback signal, TS-NODE attains substantially superior results, highlighting the critical role of the feedback loop.

\begin{table}[ht]
    \small
    \centering
    \caption{Result on LK dataset of the Baseline, TS-NODE, standard NODE trained with different augmentations, and the student in a TS-NODE without feedback. }
    \setlength{\tabcolsep}{0.53mm}{
    \begin{tabular}{cccccccc}
         \toprule
          & \multirow{2}{*}{Models} & \multirow{2}{*}{$\text{Local Error}^\downarrow$} & \multicolumn{5}{c}{$\text{Rollouts Error}^\downarrow$}\\
          &  &  &5\% & 10\% & 20\% & 50\% & 100\% \\
         \midrule
          & Baseline  & 20.93 & 0.10 & 2.54 & 64.79 & 1200.50 & 7346.09 \\
          \midrule
          & TS-NODE & \textbf{12.84} & \textbf{0.02} & \textbf{0.14} & \textbf{0.55} & \textbf{20.58} & \textbf{244.88} \\
          \midrule
          & White Noise & 23.78 & 0.05 & 0.19 & 1.02 & 42.06 & 306.50 \\
          \midrule
          & Re-scale & 106.97 & 0.01 & 1.09 & 158.53 & 4398.57 & 31925.17 \\
          \midrule
          & No Feedback & 23.60 & 0.06 & 0.23 & 2.00 & 192.03 & 2057.37 \\
          \bottomrule
    \end{tabular}
    }
    \label{tab:table_ablation}
\end{table}

\section{Related Works}
Neural ODEs \cite{node} (NODEs) show great potential in modeling complex, unknown dynamical systems. However, they can be challenging to train effectively, particularly when data is limited. Subsequent research works\cite{anode, howtotrainnode, steer} attempt to enhance the generalization of NODEs by introducing augmented dimensions, kinetic regularization, or temporal perturbations. These approaches aim to refine the structure of NODEs without directly addressing the issue of data scarcity. Parallel line of works \cite{Raissi_2018, hnn} tackle the problem by incorporating prior physics knowledge, e.g., symmetry, into modeling of dynamical systems with neural networks.

Overall, semi-supervised learning has not been thoroughly explored for dynamical system modeling. 
Data augmentation \cite{simclr, Cubuk_2019_CVPR, NEURIPS2020_d85b63ef} are not fully appropriate in this context as they can disrupt the underlying dynamical information. Among all recent works, \cite{meta_pseudo_label} most closely resembles TS-NODE. However, \cite{meta_pseudo_label} focuses on classification tasks where unlabeled data are readily available, while TS-NODE meaningfully generates and utilizes unlabeled data for dynamical systems. This contextual awareness is crucial for effective use of unlabeled data for system modeling tasks. 

\section{Limitations and Future Work}

Currently, the pseudo rollouts are simply generated around the training trajectory in a suboptimal manner, i.e., we do not optimize the locations to generate pseudo rollouts during training. An intriguing future direction is to incorporate uncertainty quantification into the framework. This could allow the model to prioritize exploration and evaluation of highly uncertain regions within the state space. By actively generating pseudo rollouts in these areas of uncertainty, TS-NODE could effectively streamline its exploration process, enhancing its understanding of the state space. Uncertainty-guided exploration may prove particularly beneficial when dealing with high-dimensional problems, where efficiently covering the state space with pseudo rollouts poses significant challenges. We leave this open problem to future exploration.


\bibliography{aaai24}

\appendix
\onecolumn
\paragraph{Quick Reference:} We briefly summarize here for better reference to specific contents. 
\begin{itemize}
    \item \cref{sec:a}: we derive the feedback gradient using a clean teacher without noise and show it is not scalable;
    \item \cref{sec:algo}: we summarize the algorithm of TS-NODE with the derived approximated feedback gradient in the main paper;
    \item \cref{sec:b}: we discuss the extension of TS-NODE framework to latent neural ODE; 
    \item \cref{sec:d}: we conduct additional experiments to report the mean and standard deviation of error metric of TS-NODE under random initializations, as well as a study on how the teacher's noise term $\sigma$ can affect the optimization process;
    \item \cref{sec:e}: we summarize detailed experimental setup and hyper-parameter settings;
    \item \cref{sec:f}: We report the computational resource and computation cost;
    \item \cref{sec:g}: We report how student perform in the main paper experiments;
\end{itemize}

\section{Unscalable Feedback Gradient with the Deterministic Teacher}\label{sec:a}
\quad The gradient of the feedback loss could be computed directly without the proposed noisy teacher. However, this approach results in considerable computational overhead, which is inefficient within the current prevalent framework of reverse-mode automatic differentiation\citep{paszke2017automatic}, as shown in the subsequent discussion.

For simplicity, we denote the shape of $\bm{\theta_{T}}$ as $|T|\times 1$ and the shape of $\bm{\theta_{S}}$ as $|S|\times 1$, respectively.
Recall that the student is updated by a one-step gradient descent based on the unlabeled loss:
\begin{equation}
    \mathcal{L}_{S}^{u} = \text{MSE}\left(\bm{y_{T}^{u}}(\bm{\theta_{T}}),\ S(\bm{y_{0}^{u}},\bm{\theta_{S}})\right)
\end{equation}
\begin{equation}
    \label{eq:s_update_app}
    \bm{\theta_{S}^{'}} = \bm{\theta_{S}} - \eta_{S}\cdot \nabla_{\bm{\theta_{S}}}\cdot \mathcal{L}_{S}^{u}
\end{equation}

The feedback loss $\mathcal{L}_{F}$ is defined as:
\begin{equation}
    \mathcal{L}_{F}^{l} := \mathcal{L}_{S^{'}}^{l}= \text{MSE}(\mathcal{D}^{l}, S^{'}(\bm{y_{0}^{l}};\bm{\theta_{S}^{'}}(\bm{\theta_{T}}))))
\end{equation}

We denote the gradient to update the student as $\bm{g_{S}}$:
\begin{equation}
    \bm{g_{S}}:=\nabla_{\bm{\theta_{S}}}\cdot \mathcal{L}_{S}^{u} = \left[\frac{\partial\mathcal{L}_{S}^{u}}{\partial\bm{\theta_{S}}}\right]^{T}
\end{equation}

We denote the gradient of the feedback loss as $\bm{g_{F}}$, by the chain rule, we have:
\begin{equation}
\begin{aligned}
    \left[\bm{g_{F}}\right]^{T}=\frac{\partial\mathcal{L}_{F}^{l}}{\partial\bm{\theta_{S^{'}}}}=\frac{\partial\mathcal{L}_{F}^{l}}{\partial\bm{\theta_{S}^{'}}}\cdot \frac{\partial\bm{\theta_{S}^{'}}}{\partial\bm{\theta_{T}}}
\end{aligned}
\end{equation}

When update the teacher's parameters, we assume the student parameters are fixed, i.e., $\bm{\theta_{T}}$ does not depend on $\bm{\theta_{S}}$. Thus, by substituting $\bm{\theta_{S}^{'}}$ in \cref{eq:s_update_app}, we have:
\begin{equation}
    \label{eq:g_f_proto}
    \begin{aligned}
    \left[\bm{g_{F}}\right]^{T}=-\eta_{S}\cdot \underbrace{\frac{\partial\mathcal{L}_{F}^{l}}{\partial\bm{\theta_{S}^{'}}}}_{a: 1\times |S|}\cdot \underbrace{\vphantom{\frac{\partial\mathcal{L}_{F}^{l}}{\partial\bm{\theta_{S}^{'}}}}\frac{\partial}{\partial\bm{\theta_{T}}}\left[\frac{\partial\mathcal{L}_{S}^{u}}{\partial\bm{\theta_{S}}}\right]^{T}}_{b: |S|\times |T|}
\end{aligned}
\end{equation}

We can further expand the unlabeled loss ($\mathcal{L}_{S}^{u}$) in \cref{eq:g_f_proto}. Let $\bm{y_{S}^{u}}(\bm{\theta_{S}})$ be the rollout of the student model from $\bm{y_{0}^{u}}$, consequently, $\mathcal{L}_{S}^{u}$ can be expressed as follows:
\begin{equation}
    \mathcal{L}_{S}^{u} = \text{MSE}\left(\bm{y_{T}^{u}}(\bm{\theta_{T}}),\ \bm{y_{S}^{u}}(\bm{\theta_{S}})\right)=\sum\limits_{i}\sum\limits_{j}\left(y_{T_{ij}}^{u}((\bm{\theta_{T}}))-y_{S_{ij}}^{u}((\bm{\theta_{S}}))\right)^{2}
\end{equation}
Where $i$ and $j$ are the time index and the dimension index of the rollout.

Now $\bm{g_{S}}$ can be expanded as:
\begin{equation}
    \label{eq:gs_expand_mse}
    \bm{g_{S}} = \left[\frac{\partial\mathcal{L}_{S}^{u}}{\partial\bm{\theta_{S}}}\right]^{T} =\sum\limits_{i}\sum\limits_{j} \left[\frac{\partial y_{S_{ij}}^{u}((\bm{\theta_{S}}))}{\partial\bm{\theta_{S}}}\right]^{T}\cdot 2\cdot \left(y_{T_{ij}}^{u}((\bm{\theta_{T}}))-y_{S_{ij}}^{u}((\bm{\theta_{S}}))\right)
\end{equation}

Then we combine \cref{eq:g_f_proto} and \cref{eq:gs_expand_mse} and derive the feedback gradient as:
\begin{equation}
    \label{eq:g_f_expand_mse}
    \left[\bm{g_{F}}\right]^{T}=-2\eta_{S}\cdot \underbrace{\frac{\partial\mathcal{L}_{F}^{l}}{\partial\bm{\theta_{S^{'}}}}}_{a: 1\times |S|}\cdot \sum\limits_{i}\sum\limits_{j} \underbrace{\vphantom{\frac{\partial\mathcal{L}_{F}^{l}}{\partial\bm{\theta_{S}^{'}}}}\left[\frac{\partial y_{S_{ij}}^{u}((\bm{\theta_{S}}))}{\partial\bm{\theta_{S}}}\right]^{T}}_{c: |S|\times 1}\cdot \underbrace{\vphantom{\frac{\partial\mathcal{L}_{F}^{l}}{\partial\bm{\theta_{S}^{'}}}}\frac{\partial y_{T_{ij}}^{u}((\bm{\theta_{T}}))}{\partial\bm{\theta_{T}}}}_{d: 1\times |T|}
\end{equation}

Unfortunately, the computation of Equation \ref{eq:g_f_expand_mse} necessitates the calculation of the gradient for each element of a rollout, as in $e$ and $d$, a process that can be prohibitively expensive. Suppose $bs$ represents the batch size, $n$ represents the number of time steps in a rollout, and $d$ denotes the dimension of the state variable. The computation of the feedback gradient would then require $O(bs\times n\times d)$ total automatic differentiation calls, implying significant computational costs.

An alternate way is to stop at \cref{eq:g_f_proto} and directly calculate $b$, i.e., we first determine $\bm{g_{S}}$ via automatic differentiation, followed by the calculation of the gradient for each parameter in $\bm{\theta_{S}}$ with respect to $\bm{\theta_{T}}$. This requires $O(|S|\times|T|)$ total numbers of calls of automatic differentiation. Meanwhile, since the computational graph needs to be maintained, this method has higher memory cost.

In summary, both methods impose substantial computational burdens and do not offer efficient implementation under the existing automatic differentiation framework. Therefore, we introduce the concept of the noisy teacher in Section 3.2, which enables us to devise a scalable and interpretable feedback gradient.

\section{Scalable TS-NODE Algorithm Flow}\label{sec:algo}
\begin{algorithm}[h]
    \footnotesize
    \caption{TS-NODE Algorithm}\label{alg:1}    
    \KwIn{Labeled data: $\mathcal{D}^{l}=\left\{\bm{y^{l}}\right\}$; Noisy teacher and student neural ODE: $T_{\epsilon}(\cdot;\bm{\theta_{T}},\sigma)$, $S(\cdot;\bm{\theta_{S}})$}
    

    \KwOut{Clean teacher: $T(\cdot;\bm{\theta_{T}^{*}})$} 
    \For{$i=0, 1, 2,\cdots N-1$}{
    1. Generate pseudo rollouts from the teacher and record the old student's performance on $\mathcal{D}^{l}$ ($\mathcal{L}_{S^{i}}^{l}$):
    
    \begin{center}
        $\bm{y_{T_{\epsilon}}^{u}} \sim \mathcal{N}\left(\bm{y_{T}^{u}};\sigma\cdot \bm{I};\bm{\theta_{T}^{i}}\right)$
    \end{center}

    2. Update the student and record the new student's performance on $\mathcal{D}^{l}$ ($\mathcal{L}_{S^{i+1}}^{l}$):
    
    \begin{center}
        $\bm{\theta_{S}^{i+1}} = \bm{\theta_{S}^{i}} - \eta_{S}\cdot \nabla_{\bm{\theta_{S}^{i}}}\cdot \mathcal{L}_{S^{i}}^{u}$
    \end{center}

    3. Calculate the feedback loss:

    \begin{center}
        $\mathcal{L}_{F}^{l}=\left(\mathcal{L}_{S^{i}}^{l}-\mathcal{L}_{S^{i+1}}^{l}\right)\cdot \text{NLL}(\bm{y}_{T_{\epsilon}}^{u};\bm{\theta_{T}^{i}})$
    \end{center}

    5. Update the teacher with its labeled loss on $\mathcal{D}^{l}$ and the feedback loss:

    \begin{center}
        $\bm{\theta_{T}^{i+1}} = \bm{\theta_{T}^{i}} - \eta_{T}\cdot \nabla_{\bm{\theta_{T}^{i}}}\cdot \left(\mathcal{L}_{T}^{l} + \mathcal{L}_{F}^{l}\right)$
    \end{center}
    }
    \Return $T(\cdot;\bm{\theta_{T}^{*}})$
\end{algorithm}
\section{Extending TS-NODE framework to Latent Neural ODEs}\label{sec:b}
\quad The proposed TS-NODE framework can be readily incorporated into latent neural Ordinary Differential Equations (ODEs)\citep{node, latent_node}, a crucial extension of neural ODEs\citep{node} designed to manage high-dimensional observations and irregularly sampled data. We refer to our framework as the latent TS-NODE and demonstrate that it can enhance extrapolation capabilities compared to conventional latent neural ODEs when dealing with limited data.
\begin{figure}[htbp!]
    \centering
    \includegraphics[width=0.85\textwidth]{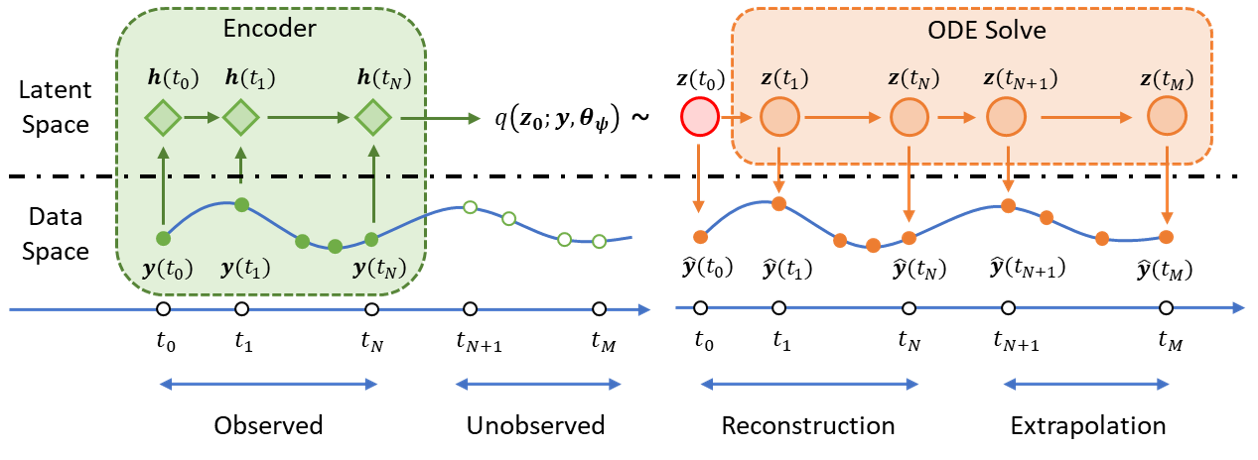}
    \caption{General framework of latent neural ODEs in \citep{latent_node}}
    \label{fig:latent_node_overview}
\end{figure}
\subsection{Basic Latent Neural ODEs}
\quad Latent neural ODEs assumes that high dimensional observations are governed by a unique dynamics in the latent space and adopt a variational autoencoder framework to model the dynamics. Given a high dimensional observation: $\bm{y}=\left\{\bm{y}(t_{0}),\bm{y}(t_{1})\cdots \bm{y}(t_{n})\right\}$ at corresponding time points $t=\left\{t_{0},t_{1}\cdots t_{n}\right\}$, latent neural ODEs models the following generative process:
\begin{itemize}
    \item [1.] Latent initial state $\bm{z_{0}}$\footnote{As in the main paper, for convenience, we write $\bm{z}(t_{0})$ as $\bm{z_{0}}$ if not otherwise specified.} is sampled from its distribution $p(\bm{z_{0}})$ in the latent space.
    \begin{equation}
        \bm{z_{0}} \sim p(\bm{z_{0}})
    \end{equation}
    \item [2.] A latent rollout is generated by solving the initial value problem with a neural ODE.
    \begin{equation}
        \hat{\bm{z}}(t_{i}) = \bm{z}(t_{0}) + \int_{t_{0}}^{t_{i}} \text{NN}(\hat{\bm{z}}(\tau), \tau;\ \bm{\theta}) d\tau.
    \end{equation}
    \item [3.] Observations at each time point ($\bm{y}(t_{i})$) are decoded independently.
    \begin{equation}
        \bm{y}(t_{i}) \sim p(\bm{y}(t_{i})|\bm{z}(t_{i})), \ i=0, 1, \cdots, n
    \end{equation}
\end{itemize}

The encoder $\psi(\cdot;\bm{\theta_{\psi}})$ in the latent neural ODEs essentially estimates the parameters for an approximated posterior distribution $q(\bm{z_{0}};\bm{y}, \bm{\theta_{\psi}})$. The decoder $\phi(\cdot;\bm{\theta_{\phi}})$ estimates the parameters for the likelihood model: $p(\bm{y}|\bm{z};\bm{\theta_{\phi}})$. An illustration of the latent neural ODEs architecture are shown in \cref{fig:latent_node_overview}.

Training of the latent neural ODEs is to maximize an evidence lower bound \citep{latent_node, kingma2022autoencoding} which balances the reconstruction error and the KL divergence between the estimated posterior $q(\bm{z_{0}}|\bm{y};\bm{\theta_{\psi}})$ and the prior distribution $p(\bm{z_{0}})$:
\begin{equation}
    \text{ELBO}(\bm{\theta_{\psi}}, \bm{\theta}, \bm{\theta_{\phi}}) = \underbrace{\mathbb{E}_{\bm{z_{0}}\sim q(\bm{z_{0}};\bm{y}, \bm{\theta_{\psi}})}\left[\log p(\bm{y}|\bm{z};\bm{\theta_{\phi}})\right]}_{-\mathcal{L}_{\text{rec}}} \underbrace{\vphantom{\mathbb{E}_{\bm{z}(t_{0})\sim q(\bm{z}(t_{0});\bm{y}, \bm{\theta_{\psi}})}\left[\log p(\bm{y}|\bm{z};\bm{\theta_{\phi}})\right]}- \mathbb{D}_{\text{KL}}\left[q(\bm{z_{0}};\bm{y}, \bm{\theta_{\psi}})||p(\bm{z_{0}})\right]}_{\mathcal{L}_{\text{KL}}}
\end{equation}

In practice, the prior, posterior and the likelihood model are all chosen to be Gaussian distributions. The encoder is usually an RNN to encode the whole sequence of observation while the decoder is a feed-forward neural network. 

\subsection{Latent TS-NODE}
\quad Latent TS-NODE introduce a student neural ODE: $S(\cdot;\bm{\theta_{S}})$ in the latent space, in addition to the existing neural ODE, which we still denote by the teacher model: $T(\cdot;\bm{\theta_{T}})$. The teacher model and the student model share the same encoder and decoder.

Latent TS-NODE operates roughly the same as the standard TS-NODE with the following steps:
\begin{itemize}
    \item [1.] Sample $\bm{z_{0}}$ from the posterior distribution $q(\bm{z_{0}}|\bm{y};\bm{\theta_{\psi}})$.
    \item [2.] The teacher generates pseudo rollouts $\bm{z_{T}^{'}}$ to train the student:
        \begin{equation}
            \mathcal{L}_{S}^{u} = \text{MSE}\left(\bm{z_{T}^{'}}(\bm{\theta_{T}}),\ S(\bm{z_{0}^{'}},\bm{\theta_{S}})\right)
        \end{equation}
        \begin{equation}
            \bm{\theta_{S}^{'}} = \bm{\theta_{S}} - \eta_{S}\cdot \nabla_{\bm{\theta_{S}}}\cdot \mathcal{L}_{S}^{u}
        \end{equation}
    \item [3.] The updated student generates latent rollout $\bm{z_{S}}$ which is decoded to $\bm{\hat{y}_{S}}$. \emph{The reconstruction error is used as the feedback loss}:
        \begin{equation}
            \mathcal{L}_{F}(\bm{\theta_{T}})\footnote{Note: in the current latent TS-NODE, $\mathcal{L}_{F}$ is used to update the  parameters of the teacher neural ODE only, it will not update the parameters of the encoder and decoder.}:=\mathcal{L}_{\text{rec}}(\bm{\hat{y}_{S^{'}}}, \bm{y}) = \mathbb{E}_{\bm{z_{0}}\sim q(\bm{z_{0}};\bm{y}, \bm{\theta_{\psi}})}\left[-\log p(\bm{y}|\bm{z_{S^{'}}};\bm{\theta_{\phi}})\right]
        \end{equation}
    \item [4.] The teacher is updated with both the ELBO and the feedback loss:
        \begin{equation}
            \bm{\theta_{T}^{'}} = \bm{\theta_{T}} - \eta_{T}\cdot \nabla_{\bm{\theta_{T}}}\left(-\text{ELBO}(\bm{\theta_{\psi}}, \bm{\theta_{T}}, \bm{\theta_{\phi}}) + \mathcal{L}_{F}\right)
        \end{equation}
\end{itemize}

For efficient computation, we can directly follow the same procedure in Section 3.2and introduce an noisy teacher ($T_{\epsilon}$) to derive the alternate feedback loss:
\begin{equation}
    \bm{z_{T_{\epsilon}}^{'}} \sim \mathcal{N}(\bm{z_{T}^{'}};\sigma\cdot \bm{I};\bm{\theta_{T}})
\end{equation}
\begin{equation}
    \begin{aligned}
        \mathcal{L}_{F} &\approx \left(\mathcal{L}_{\text{rec}}(\bm{\hat{y}_{S}}, \bm{y})-\mathcal{L}_{\text{rec}}(\bm{\hat{y}_{S^{'}}}, \bm{y})\right)\cdot \text{NLL}(\bm{z}_{T_{\epsilon}}^{'};\bm{\theta_{T}})
    \end{aligned}
\end{equation}

\begin{figure}[htb]
    \centering
    \includegraphics[width=0.47\textwidth]{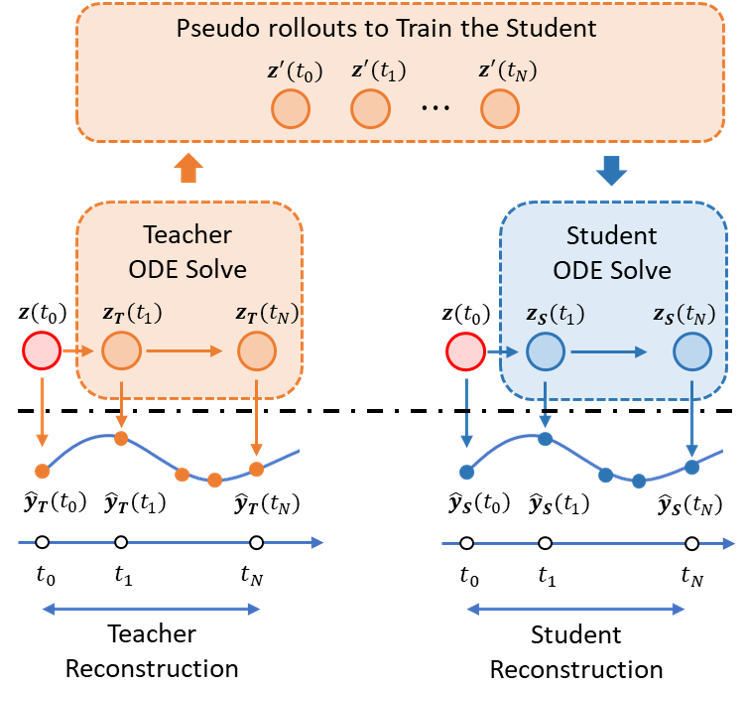}
    \caption{illustration of the additional student neural ODE and data flow in latent TS-NODE.}
    \label{fig:latent_ts_node_overview}
\end{figure}

We provide an illustration of the different parts between latent TS-NODE and the latent neural ODEs as in \cref{fig:latent_ts_node_overview}.

\subsection{Experiment}
\paragraph{Dataset:} We construct the 14 dimension Hopper dataset\citep{tunyasuvunakool2020} following the same procedure as in \citep{latent_node}. We simulate 1000 trajectories of 200 time points and use randomly selected 100 trajectories for training and the rest 900 trajectories for testing, different from \citep{latent_node}, which simulates 10000 trajectories and use 8000 for training, since we focus on a situation when observations are limited.
\paragraph{Training and Evaluation:} We use the first 100 time points in the training trajectories as observed data. We evaluate the trained model with reconstruction error and extrapolation error. 

To calculate the reconstruction error, we let the model observe the data at the first 100 time points in testing trajectories and perform reconstruction at the same time points. The MSE between the reconstructed trajectories and the observed data is defined as the reconstruction error. 

For the extrapolation error, we let the model observe the data at the first 100 time points in testing trajectories and extrapolate beyond the observed time points with different length of 5, 10, 20, 50 and 100 time points, respectively. The extrapolation error of a particular length is defined as the MSE between the extrapolated trajectories and the testing trajectories with that length. 

\paragraph{Model and Baseline:} We compare latent TS-NODE with a baseline latent neural ODE with a GRU-RNN as encoder\citep{latent_node}. For both models, we use the same architecture for the encoder, neural ODEs and the decoder. Detailed setup can be found in \cref{sec:e}.

\paragraph{Results:} The baseline and the latent TS-NODE model are trained with Adam optimizer for 10000 iterations. We record the error metrics of the two models once every 100 iterations.  \cref{tab:table_comparison_latent} reports the performances of the two models based on the averages of the last five recorded errors. 

\begin{table*}[t]
    \small
    \centering
    \setlength{\tabcolsep}{3.2mm}{
    \begin{tabular}{ccccccc}
         \toprule
          \multirow{2}{*}{Models} & \multirow{2}{*}{$\text{Reconstruction}^\downarrow$} & \multicolumn{5}{c}{$\text{Extrapolation}^\downarrow$}\\
          &  & 5 steps & 10 steps & 20 steps & 50 steps & 100 steps \\
         \midrule
          Baseline & \textbf{12.19} & 7.29 & 7.25 & 7.16 & 7.06 & 6.62  \\
          Latent TS-NODE & 12.31 & \textbf{6.06} & \textbf{5.95} & \textbf{5.79} & \textbf{5.49} & \textbf{5.13} \\
          \bottomrule
    \end{tabular}
    }
    \caption{The reconstruction error and extrapolation error of the baseline model and the latent TS-NODE model on Hopper dataset. Better results are highlighted in bold.}
    \label{tab:table_comparison_latent}
\end{table*}


\section{Additional Experiments}\label{sec:d}
\subsection{TS-NODE Under Random Initialization}\label{sec:random}
\quad For each dataset, we do five random runs and report the mean and standard deviation of the local error and the rollout error. The results are summarized in \cref{tab:table_seed}. TS-NODE can consistently surpass the baseline model in the reported runs. 
\begin{table}[ht]
    \small
    \centering
    \setlength{\tabcolsep}{2.8mm}{
    \begin{tabular}{cccccccc}
         \toprule
          \multirow{2}{*}{Dataset} & \multirow{2}{*}{Models} & \multirow{2}{*}{$\text{Local Error}^\downarrow$} & \multicolumn{5}{c}{$\text{Rollouts Error}^\downarrow$}\\
          &   &  &5\% & 10\% & 20\% & 50\% & 100\% \\
          \midrule
          \multirow{2}{*}{Cubic}& Baseline & 58.69$\pm$10.65 & 1.01$\pm$0.05 & 1.39$\pm$0.04 & 1.26$\pm$0.05 & 1.03$\pm$0.23 & 0.85$\pm$0.15  \\
          & TS-NODE & 49.76$\pm$10.70 & 0.85$\pm$0.07 & 1.21$\pm$0.11 & 1.08$\pm$0.13 & 0.81$\pm$0.18 & 0.71$\pm$0.14   \\
          \midrule
          \multirow{2}{*}{LK} & Baseline & 17.58$\pm$4.92 & 0.05$\pm$0.03 & 0.67$\pm$0.79 & 15.43$\pm$20.07 & 451.46$\pm$473.03 & 3547.28$\pm$3078.85 \\
          & TS-NODE & 13.89$\pm$1.71 & 0.02$\pm$0.01 & 0.14$\pm$0.06 & 1.97$\pm$2.21 & 88.75$\pm$122.29 & 847.83$\pm$1054.91   \\
          \midrule
          \multirow{2}{*}{Pendulum} & Baseline & 39.82$\pm$7.24 & 0.05$\pm$0.01 & 0.18$\pm$0.04 & 0.45$\pm$0.17 & 35.06$\pm$52.65 & 1088.44$\pm$1042.66  \\
          & TS-NODE & 29.31$\pm$17.77 & 0.04$\pm$0.02 & 0.14$\pm$0.07 & 0.45$\pm$0.18 & 1.13$\pm$0.35 & 18.17$\pm$31.86  \\
          \bottomrule
    \end{tabular}
    }
    \caption{Error metric comparison between the Baseline model and TS-NODE on three datasets under 5 random initialization.}
    \label{tab:table_seed}
\end{table}

\subsection{The Effect of $\sigma$ in the Noisy Teacher}
\quad In the learning process of TS-NODE, $\sigma$ is a pivotal parameter that governs the interplay between the feedback quality from the student and the teacher's adjustment scope.
\begin{itemize}
    \item Small $\sigma$: When is set to a smaller value, the pseudo rollouts tend to align closely with the teacher's mean predictions, causing the student to largely mirror the teacher and the feedback from the student reflects the teacher's knowledge more accurately. While the feedback is more precise, it remains confined to a local context.
    \item Large $\sigma$: On the other hand, a larger introduces more variance into the pseudo rollouts. Feedback from the student, which now learns from these noisier rollouts, becomes less accurate but encompasses a broader perspective. This essentially enables the teacher to probe a more expansive region around its mean prediction and adjust in a wider range.
\end{itemize}

We've conducted experiments with varying $\sigma$ values using the LK dataset, with result presented in \cref{tab:table_sigma}. As observed, an increase $\sigma$ in grants the teacher a broader scope of prediction adjustment, leading to noticeable improvements in both local error and rollout error. Yet, it's worth noting that an excessively large $\sigma$ can be counterproductive. When $\sigma$ is set too large, the feedback's inherent noise escalates, causing a rise in local errors and surpassing even the baseline. This indicates that while the feedback still exerts a regularizing effect (most likely a random noise to the gradient) and smooth the system's dynamics, the amplified noise and erroneous feedback data begin to adversely influence the teacher's learning.
\begin{table}[!ht]
    \small
    \centering
    \caption{Ablation study on the effect of different $\sigma$ on the learning of TS-NODE with LK dataset.}
    \setlength{\tabcolsep}{5.7mm}{
    \begin{tabular}{cccccccc}
         \toprule
          & \multirow{2}{*}{Models} & \multirow{2}{*}{$\text{Local Error}^\downarrow$} & \multicolumn{5}{c}{$\text{Rollouts Error}^\downarrow$}\\
          &  &  &5\% & 10\% & 20\% & 50\% & 100\% \\
         \midrule
          & Baseline  & 20.93 & 0.10 & 2.54 & 64.79 & 1200.50 & 7346.09 \\
          \midrule
          & TS-NODE ($\sigma=0.001$) & 13.38 & 0.03 & 0.17 & 1.85 & 65.95 & 708.19 \\
          \midrule
          & TS-NODE ($\sigma=0.010$) & 12.44 & 0.03 & 0.17 & 1.22 & 50.92 & 497.35 \\
          \midrule
          & TS-NODE ($\sigma=0.100$) & 12.84 & 0.02 & 0.14 & 0.55 & 20.58 & 244.88 \\
          \midrule
          & TS-NODE ($\sigma=0.368$) & 26.26 & 0.02 & 0.14 & 0.33 & 4.55 & 61.70 \\
          \midrule
          & TS-NODE ($\sigma=2.718$) & 46.10 & 0.02 & 0.16 & 0.35 & 3.67 & 65.98 \\
          \bottomrule
    \end{tabular}
    }
    \label{tab:table_sigma}
\end{table}

\section{Detailed Experimental Setup and Hyper-parameter Setting}\label{sec:e}
\subsection{Synthetic Datasets}
\quad For all synthetic datasets, we use the same architecture of a feed-forward neural network with one hidden layer for all neural ODEs in the baseline model and TS-NODE model. All models are trained with Adam optimizer with a learning rate of 0.002. Especially, the teacher and the baseline model are pre-trained for 200 iterations as warm-up. The training data are batched short observations as adopted in \citep{node} with an length of 10 time points and batch size of 50. The pseudo rollouts are generated from initial states which are sample from training trajectory and then added noise, with an std of 0.1. The batch size of pseudo rollouts are 200. 

We do not use adjoint method to solve the ODEs and utilize the default numerical solver (dopri5). Other configurations vary across datasets which we summarize independently. 
\begin{table}[htbp]
    \centering
    \setlength{\tabcolsep}{9.6mm}{
    \begin{tabular}{ccccc}
        \toprule
         Dataset & Hyper-parameters &  Baseline & Teacher & Student \\
         \multirow{2}{*}{Cubic}& Hidden units in NODE &  256 & 256 & 256 \\
         & Teacher $\sigma$ & $\backslash$ & 0.005 & $\backslash$ \\
         \midrule
         \multirow{2}{*}{LK}& Hidden units in NODE &  256 & 256 & 256 \\
         & Teacher $\sigma$ & $\backslash$ & 0.1 & $\backslash$ \\
         \midrule
         \multirow{2}{*}{Pendulum}& Hidden units in NODE &  256 & 256 & 256 \\
         & Teacher $\sigma$ & $\backslash$ & 0.005 & $\backslash$ \\
         \bottomrule
    \end{tabular}
    }
    \caption{Hyper-parameters for different synthetic datasets.}
    \label{tab:setting_synthetic}
\end{table}
\subsection{Latent TS-NODE Experiments}
\quad In the latent TS-NODE experiment, the prior distribution is a normal distribution with $\bm{0}$ mean and unit variance. The posterior distribution is a conditional Gaussian distribution with diagonal covariance matrix. The likelihood model is a conditional Gaussian distribution with a fixed diagonal covariance matrix as $0.001\cdot \bm{I}$.
we set the hidden units in the GRU cells to be 30, the last hidden state from the GRU-RNN is mapped to the mean and log standard deviation of the posterior distribution by an MLP with one hidden layer and 512 hidden units. The latent dimension is set to 15. The neural ODEs in the baseline model and the latent TS-NODE model are MLPs with 2 hidden layers with 512 hidden units each. The standard deviation for the noisy teacher model is 0.001. The decoder network is an MLP with one hidden layer and 512 hidden units.

All models are trained with Adam optimizer with a learning rate of 0.002.  
\subsection{Additional Experiments}
\quad The first additional experiment in \cref{sec:random} has exactly the setup as the main paper, the only difference is that we do not fix the random seed. The second additional experiment use the same LK dataset and NODE architecture as in \cref{tab:setting_synthetic}.

\section{Computational Resource}\label{sec:f}
\quad All experiments are run on a server with a AMD EPYC Milan 7413 CPU and with a NVIDIA A100 Ampere 80GB Graphics Card. Training of TS-NODE on single dataset with the configuration in the paper took between 40 minutes to 1 hour to complete with the aforementioned platform.

\section{Detailed Results in the Main Paper}\label{sec:g}
\begin{figure}[!htbp]
    \centering
    \includegraphics[width=0.85\textwidth]{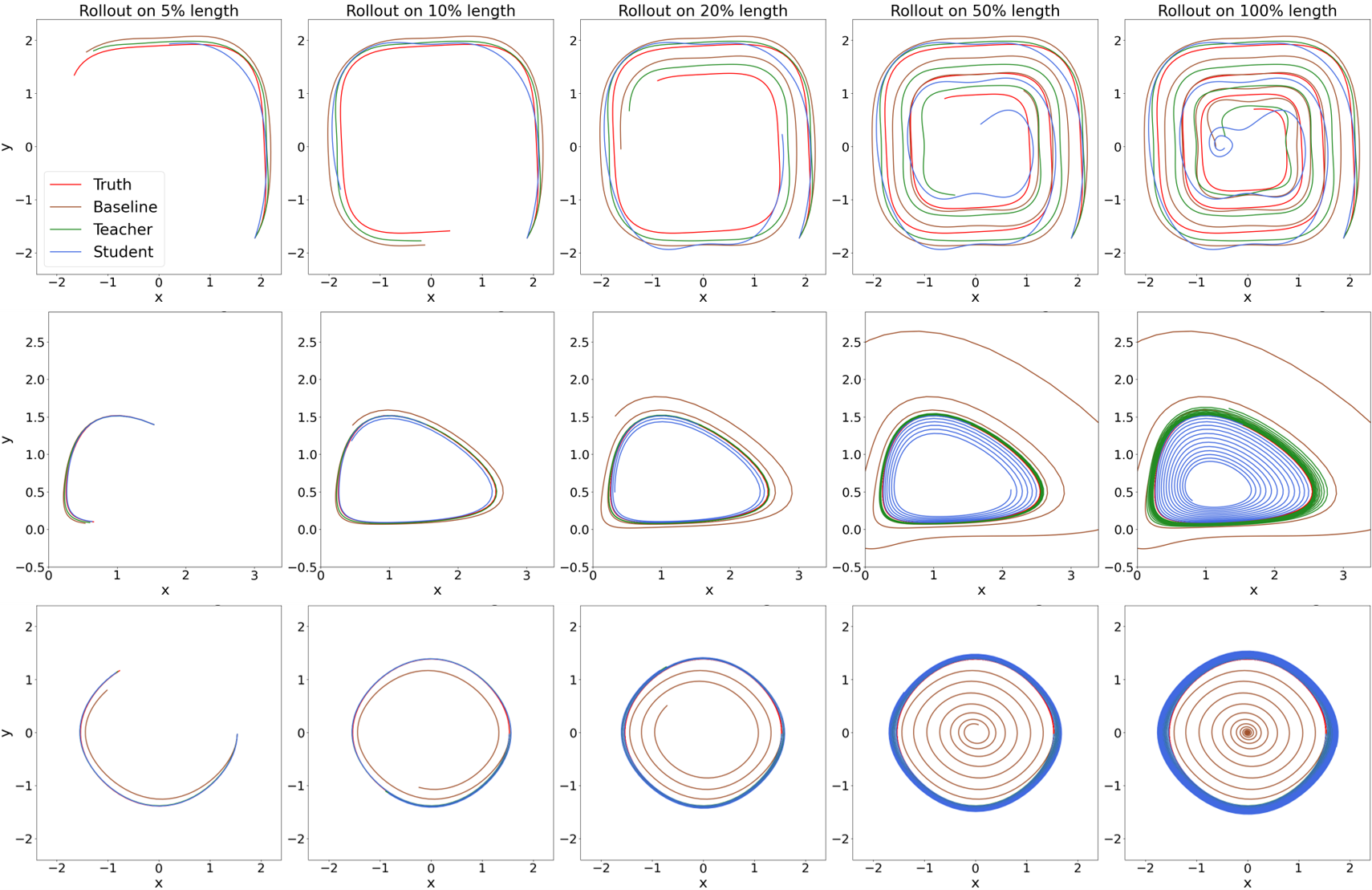}
    \caption{Rollouts from the trained baseline model and teacher model of TS-NODE in the phase space for three different systems. The rollouts are simulated from the initial condition of one pre-generated full testing trajectory for 5\%, 10\%, 20\%, 50\%, and 100\%, respectively of the full length (1,000 steps). Top: converging cubic system. Middle: Lotka-Volterra system. Bottom: pendulum. Student model included.}
    \label{fig:trace_full}
\end{figure}

\begin{figure}[!htbp]
    \centering
    \includegraphics[width=0.85\textwidth]{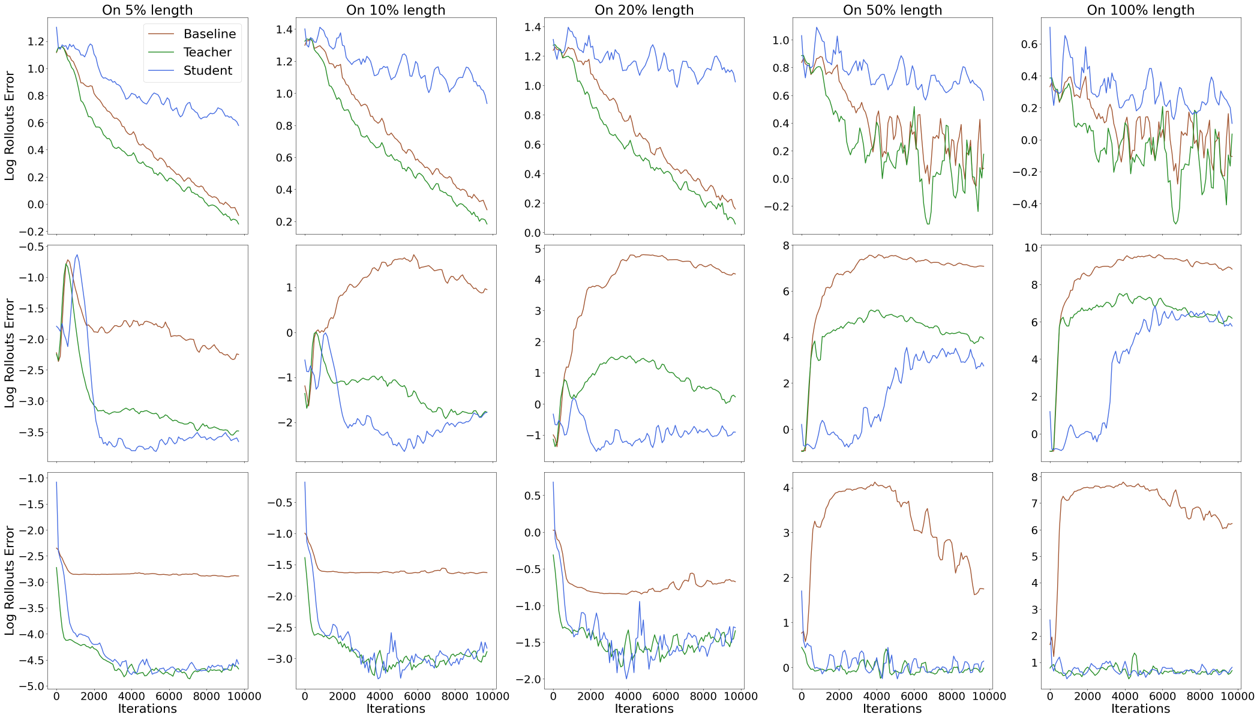}
    \caption{The rollouts errors of different length (5\%, 10\%, 20\%, 50\%, and 100\% of the full length (1000 steps)) of the trained baseline model and the teacher model of TS-NODE. Top: converging cubic system. Middle: Lotka-Volterra system. Bottom: simple pendulum. Student model included.}
    \label{fig:test_score_full}
\end{figure}

\quad In the main paper, we do not include the student error and plots for readability, we include the results with the student model in \cref{tab:table_comparison_full}. The corresponding trace plot and the test score plot are shown in \cref{fig:trace_full} and \cref{fig:test_score_full} respectively.
\begin{table*}[!hp]
    \small
    \centering
    \setlength{\tabcolsep}{6.0mm}{
    \begin{tabular}{cccccccc}
         \toprule
          \multirow{2}{*}{Dataset} & \multirow{2}{*}{Models} & \multirow{2}{*}{$\text{Local Error}^\downarrow$} & \multicolumn{5}{c}{$\text{Rollouts Error}^\downarrow$}\\
          &   &  &5\% & 10\% & 20\% & 50\% & 100\% \\
         \midrule
          \multirow{2}{*}{Cubic}& Baseline & 39.84 & 0.95 & 1.35 & 1.21 & 1.42 & 1.08 \\
          & TS-NODE & \textbf{32.77} & \textbf{0.87} & \textbf{1.21} & \textbf{1.07} & \textbf{1.12} & \textbf{0.96} \\
          & Student & 39.84 & 1.83 & 2.68 & 2.94 & 1.87 & 1.22 \\
          \midrule
          \multirow{2}{*}{LK}& Baseline & 20.93 & 0.10 & 2.54 & 64.79 & 1200.50 & 7346.09 \\
          & TS-NODE & \textbf{12.84} & 0.02 & 0.14 & 0.55 & 20.58 & 244.88 \\
          & Student & 16.08 & \textbf{0.01} & \textbf{0.07} & \textbf{0.21} & \textbf{1.69} & \textbf{51.41} \\
          \midrule
          \multirow{2}{*}{Pendulum} & Baseline & 40.49 & 0.06 & 0.20 & 0.51 & 5.77 & 507.17  \\
          & TS-NODE & \textbf{5.65} & \textbf{0.01} & \textbf{0.05} & \textbf{0.25} & \textbf{0.95} & \textbf{1.89} \\
          & Student & 40.49 & 0.01 & 0.06 & 0.27 & 1.07 & 2.04 \\
          \bottomrule
    \end{tabular}
    }
    \caption{Local error (lower the better) and rollouts error (lower the better) of the trained baseline model and the TS-NODE model on three dynamical systems. Student model included.}
    \label{tab:table_comparison_full}
\end{table*}

\end{document}